\documentclass[conference]{IEEEtran}

\IEEEoverridecommandlockouts
\usepackage{cite}
\usepackage{amsmath,amssymb,amsfonts}
\usepackage{graphicx}
\usepackage{textcomp}
\usepackage{rotating}
\usepackage{xcolor}
\usepackage{booktabs}
\usepackage{hyperref}
\usepackage{tikz}
\usetikzlibrary{arrows.meta,positioning}
\usepackage{comment}
\usetikzlibrary{positioning,calc,patterns}
\hypersetup{hidelinks}
\usepackage{layouts}
\usepackage{algorithm}
\usepackage{graphicx}
\usepackage{algpseudocode}
\usepackage{placeins}

\def\BibTeX{{\rm B\kern-.05em{\sc i\kern-.025em b}\kern-.08em
    T\kern-.1667em\lower.7ex\hbox{E}\kern-.125emX}}

\usepackage{booktabs}
\usepackage{lipsum}  
\usepackage{subcaption}

\usepackage{ulem} 

\usepackage{siunitx}

\usepackage{color}
\usepackage{tikz}
\usepackage{pgfplots}
\pgfplotsset{compat=1.18}
\usepgfplotslibrary{dateplot}
\usepgfplotslibrary{fillbetween}
\usetikzlibrary{patterns}
\usepackage{pgfplotstable}
\usepackage{csvsimple}

\usepackage{microtype}
\usepackage{graphicx}
\usepackage{booktabs} 
\usepackage{hyperref}
\usepackage{url}
\usepackage{times}
\usepackage{epsfig}
\usepackage{graphicx}
\usepackage{amsmath}
\usepackage{amssymb}
\usepackage{amsfonts}
\usepackage{dutchcal}

\usepackage{xcolor}
\usepackage{multirow}
\usepackage{siunitx}
\usepackage{textcomp}
\usepackage{braket}
\usepackage{stackengine}

\usepackage[font=footnotesize,labelfont=bf]{caption}
\usepackage{comment}
\usepackage{mathtools}

\usepackage{microtype}

\def\-{\raisebox{.75pt}{-}}

\newcommand{\vect}{\mathbf}

\usepackage{verbatim}
\usepackage{pgfplots}
\usetikzlibrary{patterns}

\usepackage{pgfplotstable}
\usepackage{colortbl}
\usepackage{graphicx}
\usepackage{booktabs}
\usepackage[export]{adjustbox}
\usepackage{siunitx}               

\usepackage{longtable}
\usepackage{tabu} 

\pgfplotstableset{
  discard if equal/.style = {
    preproc cell content/.code={
      \ifdim##1pt=#1pt
        \pgfkeyssetvalue{/pgfplots/table/@cell content}{}
      \fi
    }
  },
  color cells/.style = {
    postproc cell content/.code={
      \pgfkeyssetvalue{/pgfplots/table/@cell content}{\cellcolor{#1}##1}
    }
  }
}

\usepgfplotslibrary{groupplots}

\usepackage{enumitem}

\newcommand{\expnum}[2]{
\ifnum#1=1 
  10^{#2} 
\else 
  #1 \! \cdot \! 10^{#2}
\fi
}

\newenvironment{customlegend}[1][]{%
  \begingroup
  \csname pgfplots@init@cleared@structures\endcsname
    \pgfplotsset{#1}%
  }{%
    \csname pgfplots@createlegend\endcsname
    \endgroup
  }%
  \def\addlegendimage{\csname pgfplots@addlegendimage\endcsname}

\usepgfplotslibrary{fillbetween}
\usetikzlibrary{intersections}
\usetikzlibrary{shapes,decorations,shadows}  
\usetikzlibrary{decorations.pathmorphing}  
\usetikzlibrary{decorations.shapes}  
\usetikzlibrary{fadings}  
\usetikzlibrary{patterns}  
\usetikzlibrary{calc}

\definecolor{mycolor1}{HTML}{ca0020}
\definecolor{mycolor2}{HTML}{f4a582}
\definecolor{mycolor3}{HTML}{58d68d}
\definecolor{mycolor4}{HTML}{92c5de}
\definecolor{mycolor5}{HTML}{0571b0}
\definecolor{mycolor6}{HTML}{a6611a}
\definecolor{mycolor7}{HTML}{dfc27d}

\definecolor{cDynTrades}{RGB}{31,119,180}   
\definecolor{cDynVol}{RGB}{255,127,14}      
\definecolor{cStatTrades}{RGB}{44,160,44}   
\definecolor{cStatVol}{RGB}{140,86,75}      
\definecolor{cSelf}{RGB}{214,39,40}         
\definecolor{cRF}{RGB}{148,103,189}         
\definecolor{cXGB}{RGB}{0,0,0}              

\usepackage{hyperref}
\hypersetup{
    colorlinks,
    linkcolor={red!80!black},
    citecolor={blue!60!black},
    urlcolor={blue!80!black},
    filecolor=magenta,      
}

\def\BibTeX{{\rm B\kern-.05em{\sc i\kern-.025em b}\kern-.08em
    T\kern-.1667em\lower.7ex\hbox{E}\kern-.125emX}}
    \makeatletter
\newcommand{\linebreakand}{%
  \end{@IEEEauthorhalign}
  \hfill\mbox{}\par
  \mbox{}\hfill\begin{@IEEEauthorhalign}
}

\begin{document}

\title{Fraud Detection in Cryptocurrency Markets with Spatio-Temporal Graph Neural Networks}

\author{
    \IEEEauthorblockN{
        Lidia Losavio\IEEEauthorrefmark{1}\textsuperscript{1}\thanks{\textsuperscript{1}Equal contribution.}, 
        Luca Persia\IEEEauthorrefmark{1}\IEEEauthorrefmark{2}\textsuperscript{1}, 
        Madan Sathe\IEEEauthorrefmark{3}, and 
        Dimosthenis Pasadakis\IEEEauthorrefmark{1}
    }
    \vspace{0.2cm} 
    
    \IEEEauthorblockA{\IEEEauthorrefmark{1}\textit{Faculty of Informatics, Universit\`{a} della Svizzera italiana}, Lugano, Switzerland}
    \IEEEauthorblockA{\IEEEauthorrefmark{2}\textit{School of Management and Law, Zurich University of Applied Sciences}, Zurich, Switzerland}
    \IEEEauthorblockA{\IEEEauthorrefmark{3}\textit{Forensics and Financial Crime, Deloitte AG}, Switzerland}
    \vspace{0.2cm} 
    \IEEEauthorblockA{
        \{lidia.anna.maria.losavio, luca.persia, dimosthenis.pasadakis\}@usi.ch, msathe@deloitte.ch
    }
}


\maketitle

\thispagestyle{plain}
\pagestyle{plain}

\begin{abstract}
Technological advancements in cryptocurrency markets have increased accessibility for investors, but concurrently exposed them to the risks of market manipulations. Existing fraud detection mechanisms typically rely on machine learning methods that treat each financial asset (i.e., token) and its related transactions independently. However, market manipulation strategies are rarely isolated events, but are rather characterized by coordination, repetition, and frequent transfers among related assets. This suggests that relational structure constitutes an integral component of the signal and can be effectively represented through graphical means. In this paper, we propose three graph construction methods that rely on aggregated hourly market data. The proposed graphs are processed by a unified spatio-temporal Graph Neural Network (GNN) architecture that combines attention-based spatial aggregation with temporal Transformer encoding. We evaluate our methodology on a real-world dataset comprised of pump-and-dump schemes in cryptocurrency markets, spanning a period of over three years. Our comparative results showcase that our graph-based models achieve significant improvements over standard machine learning baselines in detecting anomalous events. Our work highlights that learned market connectivity provides substantial gains for detecting coordinated market manipulation schemes.
\end{abstract}

\begin{IEEEkeywords}
spatio-temporal Graph Neural Network, market manipulation, pump-and-dump, cryptocurrency, fraud detection
\end{IEEEkeywords}

\section{Introduction and Related Work}
\label{sec:intro}


Financial market manipulation is the attempt to distort the process by which the prices of financial assets are formed, so that they no longer reflect real information about supply and demand~\cite{AFM2017MarketManipulation}. This encompasses many behaviors, but a useful research taxonomy distinguishes between trading-based, information-based and action-based  manipulation~\cite{putnins2012}.


These phenomena can be traced back to the stock market itself~\cite{putnins2012}, yet there has recently been a notable surge in manipulative behaviour in the cryptocurrency markets. The fragmented liquidity and less stringent regulation of these markets may amplify the impact of manipulative activity, as discussed in~\cite{HamrickEtAl2021}. Notable recent cases~\cite{ShifflettVigna2018, MerewhuadervSafeMoon2022} highlight that such fraudulent activities have a detrimental effect on both investors and on the credibility of the blockchain as an innovative system.


The coordination of market manipulation in crypto markets often occurs on encrypted messaging platforms, where channels are established with the objective of attracting speculators~\cite{XuLivshits2019}. 
Empirically, these "campaigns" may leave distinct footprints in the price and the traded volume of the tokens. At the event level one typically observes: (i) a sudden increase in trading volume and number of trades over a short time interval, (ii) heavy concentration of activity on a single exchange or venue, and (iii) potential clustering of events in time, as the same groups may run multiple campaigns in quick succession. The existence of these distinct patterns in cryptocurrency market manipulation motivates a statistical learning approach for their early detection.

The current state-of-the-art methods for detecting financial market frauds exploit the underlying structure of financial interactions~\cite{Akoglu15}. They leverage graphical representations of financial data and deep learning, particularly graph neural networks (GNNs)~\cite{ma2021comprehensive}, to capture relational patterns in the network~\cite{DouEtAl2020,Schmidt24}. For cryptocurrencies in particular, fraud detection is increasingly graph-oriented, learning directly from interaction networks rather than just price or volume anomalies. In~\cite{Wu2024TokenScout,Wu2025ProfitDeceit, lin2026detecting} transaction-level cryptocurrency information is exploited (e.g., sender-receiver directed graphs), which inherently implies a direct notion of the graphical structure of the problem. However, many accessible real-world datasets provide only aggregated market observations at the token level, such as OHLCV information (open, high, low, close prices and trading volume) over fixed time intervals, number of trades, and derived indicators. In this context, the relational structure between activities is not directly observed, as there are no explicit edges or direct notions of relational interactions. Consequently, a graph-based approach requires  constructing an adjacency structure from aggregated time series, with different constructions leading to different inductive biases in the downstream GNN models.

\subsection*{Contributions and outline}

We introduce a framework that enables Spatio-Temporal GNN (ST-GNN) modeling through specific graph construction methods, utilizing only aggregated market data. To strictly evaluate and compare these construction strategies, we employ a unified ST-GNN predictor across all experiments. This architecture processes data in two stages: first, for each timestamp in a lookback window, a Transformer-style spatial GNN computes node embeddings based on the constructed graph. Second, these spatial embeddings are stacked into a sequence, enriched with positional encodings, and processed by a temporal Transformer encoder to map the spatiotemporal context to a final event probability. For the graph construction, we follow a correlation-based construction approach~\cite{daCosta2023Anomaly} that quantifies co-movement in market aggregates during the training period. We adapt this idea to the problem of fraud detection in cryptocurrency markets, and derive the adjacency matrix from correlation coefficients, which we then sparsify via set thresholds. This graph construction approach is extended beyond a single static topology with two dynamic variants. Additionally, we consider a self-adaptive graph construction variant that learns latent dependencies directly from the data, capturing interactions beyond feature-based correlations.


Our approach is evaluated on a real-world dataset of pump-and-dump (P\&D) events extracted from \cite{la2020pump} detected in the period 2017-2021. The three ST-GNN variants are compared against the widely used decision-tree methods Random Forests~\cite{breiman2001random} and XGBoost~\cite{chen2016xgboost}. Unlike existing P\&D detection datasets, where the learning problem is formed by extracting a local window around each pump event and concatenating these labeled segments~\cite{XuLivshits2019, la2020pump}, we consider a global historical window that is shared across all tokens in the sample. While local windows are well-suited to measuring event signatures, they also condition the sample on being near a pump, thereby discarding most daily market variation. This can reduce background market movements and may lead to an optimistic assessment of detection difficulty, i.e., a form of selection bias in the evaluation that results in high accuracy metrics. Our global window approach shifts the task toward a more realistic surveillance setting, i.e., a highly imbalanced learning problem in which fraudulent episodes are rare relative to the full stream of normal market activity. In particular, in our constructed dataset, we observe $\sim 0.016\%$ of anomalous events in a total of $\sim 1.9 \times 10^6$ observations.

The remainder of this paper is organized as follows. In Section~\ref{sec:background}, we briefly recap the problem of detecting fraudulent events when assuming aggregated market data samples, and present a high level overview of our approach. In Section~\ref{sec:methodology} we focus on our method for estimating graphs from these samples, and using them as input in a ST-GNN architecture to detect anomalous events. In Section~\ref{sec:Num_Res}, we perform numerical experiments on real-world data and compare with state-of-the-art decision-tree methods in order to validate our proposed routines, and finally in Section~\ref{sec:concl} we draw conclusions from this work. Our code and data are publicly available at \url{https://github.com/lidialosavio-dotcom/crypto_fraud_stGNN}.


\subsection*{Notation}

In what follows, we denote scalar quantities with lowercase, vectors with lowercase bold, and matrices with uppercase bold characters. The $(i,j)$th entry of a matrix $\vect{A}$ is symbolized by $A_{ij}$ and all entries in row $i$ or column $j$ by $\vect{A}_{i:}$ and $\vect{A}_{:j}$, respectively. 
Sets are denoted by capital calligraphic characters, for example, $\mathcal{A}$, the identity matrix as $\vect{I}$ and the vector of all ones as $\vect{e}$.

\section{Fraud Detection with Aggregated Market Data}
\label{sec:background}

We address the detection of cryptocurrency fraud events at the token
level, with hourly resolution using aggregated market data. In~\ref{sec:problem} 
we formally introduce the problem, and in~\ref{sec:GNNs} its formulation within a Graph Neural
Network (GNN) architecture. 

\subsection{Problem setting}
\label{sec:problem}
Let $\mathcal{V}=\{1,\dots,N\}$ be the set of traded tokens and let $t\in\{1,\dots,T\}$ be
the index of discrete hourly timestamps. For each token $i\in\mathcal{V}$ at time $t$, we
observe a vector of aggregate features $\mathbf{x}_{i,t}\in\mathbb{R}^{F}$ and a binary
label $y_{i,t}\in\{0,1\}$ indicating whether a fraudulent episode occurs for token
$i$ at time $t$. Let $\mathbf{X}_t\in\mathbb{R}^{N\times F}$ be the matrix
containing all the feature vectors of the tokens at time $t$. Our goal
is to learn a predictor $f_\theta$ that estimates the probability $\hat{p}_{i,t}$
of fraud for a specific token $i$ at time $t$,
\begin{equation}
\begin{aligned}
\hat{p}_{i,t} &= f_\theta\!\left(\mathbf{x}_{i,t}, \mathbf{x}_{i,t-1}, \dots, \mathbf{x}_{i,t-W+1}\right)\in[0,1],\\
\hat{y}_{i,t} &= \mathbb{I}\{\hat{p}_{i,t}\ge \gamma\},
\end{aligned}
\end{equation}
where $W$ is the lookback window and $\gamma\in(0,1)$ is a decision threshold to
determine the binary prediction $\hat{y}_{i,t}$. In
practice, $\mathbf{x}_{i,t}$ typically includes OHLCV
variables, trade counts, and additional engineered features, so that time-series 
information is available in tabular form.



\subsection{Formulation within a GNN architecture}
\label{sec:GNNs}

Graph Neural Networks (GNNs) are architectures designed to learn from graph-structured data by iteratively
updating node representations through information exchange along edges
(i.e., message passing)~\cite{corso2024graph}. Let $\mathbf{h}_i^{(k)}\in\mathbb{R}^{D}$ denote the embedding
of node $i$ after the $k$-th propagation step, with $\mathbf{h}_i^{(0)}$ typically initialized from the input features, i.e., $\mathbf{h}_i^{(0)} = \mathbf{x}_{i,t}$. A generic update then reads
\begin{equation}
\label{eq:GNN_upd}
\mathbf{h}_i^{(k)} = \mathrm{U}^{(k)}\!\left(\mathbf{h}_i^{(k-1)}, \mathrm{AGG}^{(k)}\!\left(\{\mathbf{h}_j^{(k-1)}: j \in \mathcal{N}(i)\}\right)\right),
\end{equation}
where $\mathcal{N}(i)$ is the neighborhood of $i$, $\mathrm{AGG}^{(k)}(\cdot)$ is a permutation-invariant
aggregating function over neighbors (e.g., sum/mean/max), and $\mathrm{U}^{(k)}(\cdot)$ is typically
implemented by a Multilayer Perceptron (MLP). The number of steps $k$ controls the propagation
depth; after $k$ steps, $\mathbf{h}_i^{(k)}$ incorporates information from nodes up to $k$ hops away.

In our setting, nodes correspond to traded tokens and edges encode inter-token relations derived from
aggregated market dynamics. A spatio-temporal GNN (ST-GNN) extends message passing to
time-indexed observations by combining a spatial graph aggregation at each time $t$ with a temporal model that
captures how node representations evolve across a
lookback window~\cite{yu2018spatio,jin2023spatio,wu2020connecting}. Many GNN variants can be interpreted
within this message-passing template, differing mainly in the way neighbor
information is computed and propagated~\cite{GCN,GAT,GIN}.

A practical constraint in cryptocurrency fraud detection arises from relying on
exchange- or aggregator-provided APIs~\cite{BinanceSpotAPI,CoinGeckoAPI} for
data collection. These sources typically expose only market-level aggregates
per token and time (e.g., OHLCV and related features) rather than
transaction-level interactions. Consequently, the relational structure between
assets remains latent, as there are no explicit edges and any notion of
connectivity must be inferred from the observed
time series~\cite{Pasadakis23a}. This limits the applicability of many graph-based approaches that
rely on a pre-defined interaction network, thus motivating the development of 
methods capable of constructing graphs from aggregated data, and utilizing them 
in the prediction task.

\section{Inference Graphs Methods and a Unified Spatio-Temporal GNN}
\label{sec:methodology}

We infer the graphical structure of token
interactions from 
aggregated data, and 
utilize it in a spatio-temporal GNN to detect 
fraudulent events.
Subsection~\ref{sec:temporal_window} defines
data representation and temporal windowing, and
in Subsection~\ref{sec:graph_inference} introduces our graph inference strategies.
Then, Subsection~\ref{sec:stgnn_arch} details the proposed GNN architecture.

\subsection{Data representation and temporal windowing}
\label{sec:temporal_window}

Given the node feature matrix $\mathbf{X}_t\in\mathbb{R}^{N\times F}$, temporal context is captured through a fixed lookback window of length $W$. For each node 
$i \in \mathcal{V}$ at time $t$, we form the sequence
\begin{equation}
\mathbf{X}^{(W)}_{i,t} =
\big[\mathbf{x}_{i,t-W+1}, \ldots, \mathbf{x}_{i,t}\big]\in\mathbb{R}^{W\times F}.
\end{equation}
Stacking these sequences for all nodes yields the windowed tensor 
$\mathbf{X}^{(W)}_t\in\mathbb{R}^{N\times W\times F}$, whose $i$-th slice satisfies
$\mathbf{X}^{(W)}_t[i,:,:] = \mathbf{X}^{(W)}_{i,t}$. The model outputs a probability
per node for the current time $t$, i.e., 
$\hat{\mathbf{p}}_t\in[0,1]^N$.


\subsection{Graph inference strategies}
\label{sec:graph_inference}

We denote by $\mathbf{A}_t\in\mathbb{R}^{N\times N}$ a weighted adjacency matrix at time $t$, and
by $\mathcal{E}_t = \{(i,j): A_{t,ij}\neq 0\}$ the corresponding directed edge set. Undirected
connections are treated as directed edges from $i$ to $j$ and vice versa. When
required for message
passing, we equivalently represent $\mathbf{A}_t$ as an edge list $\mathcal{E}_t$ with associated
edge weights $\mathbf{a}_t=\{A_{t,ij}:(i,j)\in\mathcal{E}_t\}$. Since the underlying market 
graph is latent, we define an inference operator
\begin{equation}
\mathbf{A}_t = g_t(\mathcal{D}_{\le t}),
\end{equation}
which constructs $\mathbf{A}_t$ relying solely on
information available up to time $t$. This
ensures that the graph construction 
respects the temporal order of observations, and in practice,
is restricted only to training timestamps when building training-time
graphs~\cite{shang2021discrete}. Here, $\mathcal{D}_{\le t}$ denotes the observed aggregated
market data up to time $t$, i.e., the time series of token-level data used to
build $\mathbf{X}_t$ and any scalar time series $s_{i,t}$ employed by the graph inference rule. We
evaluate three graph construction methods for $g_t$, and summarize them in
Algorithm~\ref{alg:graph_inference}.


\begin{algorithm}[!htb]
\caption{Graph inference strategies from aggregated data}
\label{alg:graph_inference}
\begin{algorithmic}[1]
\Require Training timestamps $\mathcal{T}_{\mathrm{tr}}$, fraud timestamps $\mathcal{P}_{\mathrm{tr}}$, scalar feature $s_{i,t}$, density $\rho$, min threshold $\tau_{\min}$, dynamic horizon $L{=}12$, self-adaptive threshold $\epsilon$
\Ensure Inferred adjacency (static $\mathbf{A}$, dynamic updates $\{\mathbf{A}^{(p)}\}$, or self-adaptive $\mathbf{A}$)
\Statex \textbf{(G1) Static}
\State Compute $\mathbf{C}$ on $\mathcal{T}_{\mathrm{tr}}$ from $\log(1+s_{i,t})$
\State $\tau \gets \max(\tau_{\min}, Q_{\rho}(\{C_{ij}\}_{i<j}))$
\State $A_{ij} \gets C_{ij}\cdot\mathbb{I}\{C_{ij}>\tau\}$
\Statex \textbf{(G2) Dynamic event-driven}
\For{each $p\in\mathcal{P}_{\mathrm{tr}}$ in chronological order}
    \State $\mathcal{W}(p)\gets \{p-L,\dots,p\}$
    \State Compute $\mathbf{C}^{(p)}$ on $\mathcal{W}(p)$
    \State $\tau^{(p)} \gets \max(\tau_{\min}, Q_{\rho}(\{C^{(p)}_{ij}\}_{i<j}))$
    \State Add/update edges $(i,j)$ where $C^{(p)}_{ij}>\tau^{(p)}$ using running mean
    \State Save $\mathbf{A}^{(p)}$
\EndFor
\Statex \textbf{(G3) Self-adaptive}
\State Learn $\mathbf{E}_1,\mathbf{E}_2\in\mathbb{R}^{N\times d}$; compute $\mathbf{M}\gets \mathbf{E}_1\mathbf{E}_2^\top$
\State $\mathbf{A}\gets \mathrm{softmax}(\mathrm{ReLU}(\mathbf{M}))$ (row-wise)
\State Sparsify: keep $(i,j)$ only if $A_{ij}>\epsilon$
\end{algorithmic}
\end{algorithm}

\subsection*{(G1) Static correlation graph}
\label{sec:static_graph}

We construct a single weighted adjacency from the training period and reuse it at every timestamp. Let
$\mathcal{T}_{\mathrm{tr}} \subseteq \mathcal{T}$
denote the set of training timestamps. Given a
scalar time series $s_{i,t}$ for each token $i$,
obtained from a single chosen node feature
(e.g., volume or number of trades), we define the
log-transformed token-by-time matrix
$\mathbf{S}\in\mathbb{R}^{|\mathcal{T}_{\mathrm{tr}}|\times N}$ with entries:
\begin{equation}
S_{t,i}=\log(1+s_{i,t}), \qquad t\in\mathcal{T}_{\mathrm{tr}},\ i\in\{1,\dots,N\}.
\end{equation}
We compute the Pearson correlation matrix $\mathbf{C}\in[-1,1]^{N\times N}$ and set the diagonal
entries to zero:
\begin{equation}
C_{ij}=\mathrm{Corr}\!\big(\mathbf{S}_{:,i},\mathbf{S}_{:,j}\big),\qquad C_{ii}=0.
\end{equation}
To sparsify this matrix, we select a data-driven
threshold $\tau$ via an off-diagonal quantile,
subject to a 
minimum floor $\tau_{\min}$:
\begin{equation}
\tau = \max\Big\{\tau_{\min},\ Q_{\rho}\big(\{C_{ij}\}_{i<j}\big)\Big\},
\qquad
A_{ij}=C_{ij}\,\mathbb{I}\{C_{ij}>\tau\}\ ,
\end{equation}
with $A_{ii}=0$ and $(i\neq j)$. Here $Q_{\rho}$
denotes the $\rho$-quantile of the off-diagonal
entries. This threshold retains only the strongest
edges, i.e., the top $1 - \rho$ fraction of
them, while $\tau_{\min}$ ensured that retained
edges represent statistically meaningfull 
correlations.

Since the resulting adjacency matrix
is time-invariant, we have
$
\mathbf{A}_t=\mathbf{A}, \, \forall \, t.
$
This approach assumes that even without explicit
interactions, tokens can
exhibit synchronized market activity. A static
correlation graph is the simplest relational
inductive bias compatible with aggregated token
features~\cite{tumminello2010correlation}.

\subsection*{(G2) Event-driven dynamic correlation graph}
\label{sec:dynamic_graph}

Static graphs often fail to capture correlation
shifts associated with manipulation episodes, as
they rely on a fixed representation
of the data, and do not
update over time~\cite{matsunaga2019exploring}. Our
second graph construction approach addresses this
limitation by constructing a piecewise-constant
adjacency matrix that updates only around fraud events within the training data.

Let $\mathcal{P}_{\mathrm{tr}}\subseteq\mathcal{T}_{\mathrm{tr}}$ be the set of training timestamps
where at least one fraud event occurs. For each $p\in\mathcal{P}_{\mathrm{tr}}$, we consider a lookback window of $L$ hours:
\begin{equation}
\mathcal{W}(p)=\{p-L,\dots,p\}\cap \mathcal{T}_{\mathrm{tr}} .
\end{equation}
Note that $L$ denotes a graph construction lookback window, distinct from the feature window $W$ in Section~\ref{sec:temporal_window}. Using the same scalar series $s_{i,t}$ as in
(G1), derived from volume or number of trades, we
compute a window-specific correlation matrix 
$\mathbf{C}^{(p)}$ from $\{\log(1+s_{i,t}) :
t\in\mathcal{W}(p)\}$, enforcing again
$C^{(p)}_{ii}=0$. We
then select edges representing significant
interactions with a
density-controlled threshold:
\begin{align}
\tau^{(p)} &= \max\Big\{\tau_{\min},\ Q_{\rho}\big(\{C^{(p)}_{ij}\}_{i<j}\big)\Big\}, \\
\qquad
\mathcal{E}^{(p)} &= \{(i,j): i<j,\ C^{(p)}_{ij}>\tau^{(p)}\}.
\end{align}
For each undirected edge $(i,j)$ we keep
a cumulative sum $S_{ij}$ of the correlation
coefficients $C^{(p)}_{ij}$, and a counter
$n_{ij}$ representing the number of fraud event windows 
where the edge $(i,j)$ was part of the graph, i.e., 
$(i,j) \in \mathcal{E}^{(p)}$. The resulting edge 
weight is defined as the mean of these correlations,
$w_{ij}=S_{ij}/n_{ij}$.
The updated adjacency matrix $\mathbf{A}^{(p)}$
is formed by setting 
$A^{(p)}_{ij} = A^{(p)}_{ji}=w_{ij}$ and
$A^{(p)}_{ij}=0$ otherwise. 

At any time $t$, the model uses the graph state in
the training set of the most recent fraud timestamp
$p^\star$,
\begin{equation}
\mathbf{A}_t=
\begin{cases}
\mathbf{I}, & \text{if } \{p\in\mathcal{P}_{\mathrm{tr}}:p\le t\}=\emptyset,\\
\mathbf{A}^{(p^\star)}, & p^\star=\max\{p\in\mathcal{P}_{\mathrm{tr}}:p\le t\}.
\end{cases}
\end{equation}
For timestamps in the validation and testing
sets, the graph $\mathbf{A}^{(p_{\mathrm{last}})}$
of the last training update is considered, with $p_{\mathrm{last}}=\max\mathcal{P}_{\mathrm{tr}}$ being the last fraud event timestamp in the training set.

\subsection*{(G3) Self-adaptive adjacency from node embeddings}
\label{sec:self_adaptive_graph}
While correlation-based graphs capture statistical synchronization between assets, they rely on
fixed heuristic thresholds to determine connectivity, and thus, may not align with the 
optimal latent graph for fraud detection. To allow our model to discover latent dependencies
directly from the data, we employ a self-adaptive adjacency construction approach inspired 
by~\cite{wu2019graph}.

We randomly initialize two learnable node embedding dictionaries 
$\mathbf{E}_1,\mathbf{E}_2\in\mathbb{R}^{N\times d}$, representing source and target node
embeddings, respectively. These embeddings are optimized together with the neural network weights
via stochastic gradient descent, guided solely by minimizing the supervised classificaiton 
objective. The hyperparameter $d$ controls the rank of the learned
adjacency matrix $\mathbf{A}$. The adjacency matrix, that can be considered as the transition
matrix of a hidden diffusion process~\cite{wu2019graph}, is then computed as
\begin{equation}
\mathbf{A} = \mathrm{SoftMax}\big(\mathrm{ReLU}(\mathbf{E}_1 \mathbf{E}_2^\top)\big).
\end{equation}
Here, the $\mathrm{ReLU}$ activation function eliminates weak or negative connections, while the
row-wise $\mathrm{SoftMax}$ normalizes the weights to obtain a row-stochastic adjacency.

To sparsify this dense graph, we set edges with weights below a small tuned constant 
$\epsilon$ to zero:
\begin{equation}
\mathcal{E} = \{(i,j): A_{ij} > \epsilon\},
\qquad
A_{ij}\leftarrow A_{ij}\mathbb{I}\{A_{ij}>\epsilon\},
\end{equation}
Unlike the undirected graphs in (G1) and (G2), this approach generates an unsymmetric matrix
$\mathbf{A} \neq \mathbf{A}^{\top}$.

\subsection{ST-GNN architecture}
\label{sec:stgnn_arch}
All inferred graphs are used within the same spatio-temporal model architecture.
At each hour $t$, the inferred adjacency $\mathbf{A}_t$ defines the neighborhood
$\mathcal{N}_t(i)=\{j: A_{t,ij}\neq 0\}$. We encode
(i) spatial interactions on $\mathbf{A}_t$ with a Transformer-style attention GNN
layer~\cite{shi2020masked}, and (ii) temporal dependencies via a Transformer encoder over
an $W$-hour lookback window~\cite{vaswani2017attention, jin2023spatio}.

First, spatial aggregation is performed at each time step. For each node 
$i\in\mathcal{V}$, the input feature vector $\mathbf{x}_{i,t}$ is mapped to a spatial
embedding $\mathbf{h}_{i,t}\in\mathbb{R}^{D}$ via attentive message passing over $\mathbf{A}_t$:
\begin{equation}
\label{eq:GTransformer}
\mathbf{h}_{i,t} = \mathrm{GraphTransformer}(\mathbf{x}_{i,t}, \mathbf{A}_t).
\end{equation}
Subsequently, we construct the temporal sequence by stacking the embeddings from the lookback
window $W$:
\begin{equation}
\mathbf{H}_{i,t} = \big[\mathbf{h}_{i,t-W+1},\dots,\mathbf{h}_{i,t}\big]\in\mathbb{R}^{W\times D}.
\end{equation}
We add learnable positional encodings into $\mathbf{H}_{i,t}$, and process the sequence with a 
temporal Transformer encoder, denoted by $\mathrm{TE}(\cdot)$.
The fraud event probability for node $i$ at time $t$ is obtained by
\begin{equation}
\hat{p}_{i,t}=\sigma\!\left(\mathbf{w}^{\top}\mathrm{TE}(\mathbf{H}_{i,t})[W]+b\right),
\end{equation}
where $\mathbf{w}\in\mathbb{R}^{D}$ and $b\in\mathbb{R}$ are learnable output parameters and
$\sigma(\cdot)$ is the sigmoid logistic function. Algorithm~\ref{alg:stgnn_forward} summarizes the
forward pass.

\begin{algorithm}[!htb]
\caption{ST-GNN forward pass}
\label{alg:stgnn_forward}
\begin{algorithmic}[1]
\Require Windowed node features $\mathbf{X}^{(W)}_t\in\mathbb{R}^{N\times W\times F}$, inferred graph $(\mathcal{E}_t, \mathbf{a}_t)$ where $\mathbf{a}_t$ are edge weights, parameters $\theta$
\Ensure Event probabilities $\hat{\mathbf{p}}_t\in[0,1]^N$
\State $\texttt{seq} \gets [\;]$ \Comment{will store spatial embeddings for each hour in the window}
\For{$u=1$ to $W$} \Comment{iterate over window steps}
    \State $\mathbf{H} \gets \mathbf{W}_{\mathrm{in}}\mathbf{X}^{(W)}_t[:,u,:]$ \Comment{linear projection}
    \State $\mathbf{H} \gets \mathrm{TransformerConv}_1(\mathbf{H}, \mathcal{E}_t, \mathbf{a}_t)$
    \State $\mathbf{H} \gets \mathrm{ReLU}(\mathbf{H})$; \;\; $\mathbf{H} \gets \mathrm{Dropout}(\mathbf{H})$
    \State $\mathbf{H} \gets \mathrm{TransformerConv}_2(\mathbf{H}, \mathcal{E}_t, \mathbf{a}_t)$
    \State $\mathbf{H} \gets \mathrm{ReLU}(\mathbf{H})$
    \State $\texttt{seq}.\mathrm{append}(\mathbf{H})$
\EndFor
\State $\mathbf{S} \gets \mathrm{Stack}(\texttt{seq}) \in \mathbb{R}^{N\times W\times D}$
\State $\mathbf{S} \gets \mathbf{S} + \mathbf{P}$ \Comment{positional embedding}
\State $\tilde{\mathbf{S}} \gets \mathrm{TemporalTransformerEncoder}(\mathbf{S})$
\State $\mathbf{z} \gets \tilde{\mathbf{S}}[:,W,:]$ \Comment{last time token}
\State $\hat{\mathbf{p}}_t \gets \sigma(\mathbf{W}_{\mathrm{out}}\mathbf{z} + \mathbf{b})$
\State \Return $\hat{\mathbf{p}}_t$
\end{algorithmic}
\end{algorithm}

\section{Numerical Results}
\label{sec:Num_Res}
In this section, we present experimental results on real-world data for an extensive evaluation of the accuracy and efficacy of the proposed graph-based fraud detection routines outlined in
Section~\ref{sec:methodology}. We will compare our methods against the following
decision-tree based ensemble methods:
\begin{enumerate}
    \item Random Forests (RF)~\cite{breiman2001random}: Aggregates predictions from decorrelated
    decision trees trained on bootstrap samples and feature subsampling. We use a grid search to
    optimize the number of estimators $n_{\mathrm{est}} \in \{500, 1000\}$), maximum tree depth
    $d_{\mathrm{max}} \in \{12, \infty\}$, minimum samples per leaf
    $n_{\mathrm{min}} \in \{1, 10\}$, and the feature subset size considered at each split
    $F_{\mathrm{sub}} \in \{\sqrt{F}, 0.8F\}$.
    \item eXtreme Gradient Boosting (XGBoost)~\cite{chen2016xgboost}: A gradient-boosted tree
    method that sequentially fits trees to correct previous errors. We tune the learning rate
    $\eta \in \{0.05, 0.1\}$, maximum tree depth ($d_{\text{max}} \in \{4, 10\}$), number of 
    estimators ($n_{\text{est}} \in \{500, 1000\}$), and L2 regularization parameter 
    $\lambda \in \{1.0, 10.0\}$, while fixing row and column subsampling rates at $0.8$.
\end{enumerate}

The following Subsection~\ref{sec:experimental_setup} summarizes our experimental setup, and then~\ref{sec:Res_ACC} is devoted in a comparison of the accuracy of the methods under consideration. 


\subsection{Experimental setup}
\label{sec:experimental_setup}

We consider a real-world dataset of pump-and-dump events from, covering the period July 2017 -- January 2021, where labels where assigned by monitoring encrypted messaging platforms and linking each market manipulation campaign to a target token~\cite{LaMorgia2020}. We collect the
available market data for quote currency BTC through the Binance API~\cite{BinanceSpotAPI}, and assign a unique timestamp by combining the date and hour fields and localizing the resulting time to UTC. Since market data are represented as hourly candles\footnote{In financial market microstructures, a candle (or candlestick) aggregates trading activity over a fixed time interval and is typically represented by the OHLC information, together with volume and related trade counts for that interval.}, pump timestamps are rounded to the nearest hour boundary to enforce a consistent alignment between the pump schedule and the candle grid. A pump episode from our dataset is illustrated in Figure~\ref{fig:pair_timeseries_trades}, where the number of trades spikes sharply around the event time for two tokens, highlighting both the market signature of coordinated activity and the presence of cross-token co-movement that motivates our graph-based approach.

For each token we download a continuous time interval spanning from the earliest to the latest pump timestamp in the schedule, both extended by a fixed margin of seven days. Consequently, all tokens share the same time boundaries, resulting in a panel that includes both normal market activity and rare fraud episodes, 
contains the candle timestamp, OHLCV information and trading
activity. The binary fraud label flag is assigned by checking whether the candle timestamp belongs to the set of pump timestamps associated with that token. 
The dataset contains 1{,}905{,}850 observations for $N=84$ tokens. The positive class (flagged pump hours) includes 314 positives, corresponding to a positive rate of $\sim 0.016\%$.

\begin{figure}
\begin{tikzpicture}
\begin{axis}[
    width=0.96\columnwidth,
    height=6cm,
    date coordinates in=x,
    tick align=outside,
    axis x line*=bottom,
    axis y line*=left,
    axis line style={-},
    ylabel={Number of Trades},
    xlabel={Time},
    xticklabel style={rotate=45, anchor=east},
    xticklabel=\month-\day,
    enlarge x limits=0.18,
    grid=none,
]
\addplot[draw=none, fill=cDynTrades, fill opacity=0.15] table[
    x=date, y=num_trades, col sep=comma
]{data/BRDBTC_subset.csv}
\closedcycle;
\addplot[thick, cDynTrades] table[
    x=date, y=num_trades, col sep=comma
]{data/BRDBTC_subset.csv};
\addplot[draw=none, fill=cSelf, fill opacity=0.20] table[
    x=date, y=num_trades, col sep=comma
]{data/OAXBTC_subset.csv}
\closedcycle;
\addplot[thick, cSelf] table[
    x=date, y=num_trades, col sep=comma
]{data/OAXBTC_subset.csv};
\end{axis}
\begin{customlegend}[
    legend columns=2,
    legend style={
      at={(current axis.north)},
      anchor=south,
      yshift=6pt,
      draw=none,
      fill=none,
      /tikz/every even column/.append style={column sep=0.15cm}
    },
    legend entries={BRD\\ OAX\\},
    legend image post style={xscale=0.8},
]
  \addlegendimage{area legend, fill=cDynTrades,  draw=black, legend image post style={xscale=0.55,yscale=2}}
  \addlegendimage{area legend, fill=cSelf, draw=black, legend image post style={xscale=0.55,yscale=2}}
\end{customlegend}
\end{tikzpicture}
\caption{Hourly number of trades for BRD and OAX over a four-day window around one pump event.} 
\label{fig:pair_timeseries_trades} 
\end{figure}
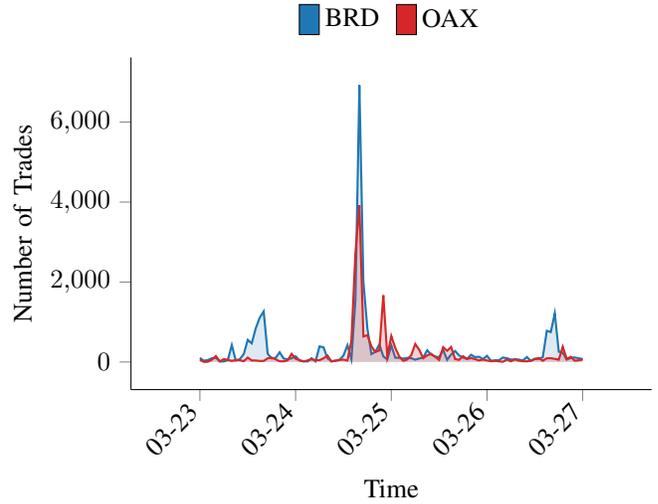


\begin{figure*}[!t]
    \centering
    \begin{minipage}{\textwidth}
      \centering
      %
  \begin{tikzpicture}
  \centering
  \begin{customlegend}[
      legend columns=7,
      legend style={
        anchor=north,
        draw=none,
        fill=none,
        /tikz/every even column/.append style={column sep=0.10cm}},
      legend entries={XGBoost\\RF\\Static (Volume)
      \\Dyn (Volume) \\Dyn (Num. Trades) \\ Static (Num. Trades)
      \\ Self-Adaptive \\},
      legend image post style={xscale=0.8},
  ]
    \addlegendimage{area legend, fill=cXGB,       draw=black, legend image post style={xscale=0.55,yscale=2.0}}
    \addlegendimage{area legend, fill=cRF,        draw=black, legend image post style={xscale=0.55,yscale=2.0}}
    \addlegendimage{area legend, fill=cStatVol,   draw=black, legend image post style={xscale=0.55,yscale=2.0}}
    \addlegendimage{area legend, fill=cDynVol,    draw=black, legend image post style={xscale=0.55,yscale=2.0}}
    \addlegendimage{area legend, fill=cDynTrades, draw=black, legend image post style={xscale=0.55,yscale=2.0}}
    \addlegendimage{area legend, fill=cStatTrades,draw=black, legend image post style={xscale=0.55,yscale=2.0}}
    \addlegendimage{area legend, fill=cSelf,      draw=black, legend image post style={xscale=0.55,yscale=2.0}}
  \end{customlegend}
\end{tikzpicture}%

      \vspace{0.2em}
    \end{minipage}
    \\
    \subcaptionbox{\label{fig:F1Bars}}{%
    \begin{minipage}{0.48\textwidth}
      \centering
      %
  \begin{tikzpicture}
\centering
\begin{axis}[
    width=0.75\linewidth,
    height=6cm,
    scale only axis,
    ybar,
    bar width=15pt,      
    bar shift=0pt,         
    tick align=outside,
    axis x line*=bottom,
    axis y line*=left,
    axis line style={-},
    ylabel={F1 score},
    xlabel={Method},
    ymin=0.00,
    ymax=0.70,
    ytick={0.1,0.2,0.3,0.4,0.5,0.6,0.7, 0.8},
    symbolic x coords={
        XGB,
        RF,
        Static (Volume),
        Dyn (Volume),
        Dyn (Num\_Trades),
        Static (Num\_Trades),
        Self-Adaptive
    },
    xtick=data,
    xticklabels={},            
    x tick style={draw=none},   
    enlarge x limits=0.10,
    error bars/y dir=both,
    error bars/y explicit,
]

\addplot[draw=black, fill=cXGB, mark=none,
    error bars/.cd, y dir=both, y explicit] coordinates {(XGB, 0.4984) +- (0, 0.0298)};

\addplot[draw=black, fill=cRF, mark=none,
    error bars/.cd, y dir=both, y explicit] coordinates {(RF, 0.5294) +- (0, 0.0143)};

\addplot[draw=black, fill=cStatVol, mark=none,
    error bars/.cd, y dir=both, y explicit] coordinates {(Static (Volume), 0.5756) +- (0, 0.0383)};

\addplot[draw=black, fill=cDynVol, mark=none,
    error bars/.cd, y dir=both, y explicit] coordinates {(Dyn (Volume), 0.5803) +- (0, 0.0285)};

\addplot[draw=black, fill=cDynTrades, mark=none,
    error bars/.cd, y dir=both, y explicit] coordinates {(Dyn (Num\_Trades), 0.5806) +- (0, 0.0373)};

\addplot[draw=black, fill=cStatTrades, mark=none,
    error bars/.cd, y dir=both, y explicit] coordinates {(Static (Num\_Trades), 0.6013) +- (0, 0.0387)};

\addplot[draw=black, fill=cSelf, mark=none,
    error bars/.cd, y dir=both, y explicit] coordinates {(Self-Adaptive, 0.6179) +- (0, 0.0475)};

\end{axis}
\end{tikzpicture}%

    \end{minipage}}%
    \hspace{0.02\textwidth}%
    \subcaptionbox{\label{fig:PRCurve}}{%
    \begin{minipage}{0.48\textwidth}
      \centering
      %
  \begin{tikzpicture}
\centering
\begin{axis}[
    width=0.75\linewidth,
    height=6cm,
    scale only axis,
    xmin=0.5, xmax=1,
    enlarge x limits=false,   
    enlargelimits=false,
    clip=true,
    xlabel={Recall},
    ylabel={Precision},
    ymin=0, ymax=0.7,
    ytick={0.1,0.2,0.3,0.4,0.5,0.6,0.7,0.8},
    tick align=outside,
    axis x line*=bottom,
    axis y line*=left,
    axis line style={-},
    grid=none,
]

\addplot[name path=HiDynTrades, draw=none] table[x=recall, y=Hi_DynNum_trades, col sep=comma]{data/mean_pr_curves_for_latex.csv};
\addplot[name path=LoDynTrades, draw=none] table[x=recall, y=Lo_DynNum_trades, col sep=comma]{data/mean_pr_curves_for_latex.csv};
\addplot[fill=cDynTrades, fill opacity=0.12, draw=none] fill between[of=LoDynTrades and HiDynTrades];
\addplot[thick, cDynTrades] table[x=recall, y=P_DynNum_trades, col sep=comma]{data/mean_pr_curves_for_latex.csv};

\addplot[name path=HiDynVol, draw=none] table[x=recall, y=Hi_DynVolume, col sep=comma]{data/mean_pr_curves_for_latex.csv};
\addplot[name path=LoDynVol, draw=none] table[x=recall, y=Lo_DynVolume, col sep=comma]{data/mean_pr_curves_for_latex.csv};
\addplot[fill=cDynVol, fill opacity=0.12, draw=none] fill between[of=LoDynVol and HiDynVol];
\addplot[thick, cDynVol] table[x=recall, y=P_DynVolume, col sep=comma]{data/mean_pr_curves_for_latex.csv};

\addplot[name path=HiStatTrades, draw=none] table[x=recall, y=Hi_StaticNum_Trades, col sep=comma]{data/mean_pr_curves_for_latex.csv};
\addplot[name path=LoStatTrades, draw=none] table[x=recall, y=Lo_StaticNum_Trades, col sep=comma]{data/mean_pr_curves_for_latex.csv};
\addplot[fill=cStatTrades, fill opacity=0.12, draw=none] fill between[of=LoStatTrades and HiStatTrades];
\addplot[thick, cStatTrades] table[x=recall, y=P_StaticNum_Trades, col sep=comma]{data/mean_pr_curves_for_latex.csv};

\addplot[name path=HiStatVol, draw=none] table[x=recall, y=Hi_StaticVolume, col sep=comma]{data/mean_pr_curves_for_latex.csv};
\addplot[name path=LoStatVol, draw=none] table[x=recall, y=Lo_StaticVolume, col sep=comma]{data/mean_pr_curves_for_latex.csv};
\addplot[fill=cStatVol, fill opacity=0.12, draw=none] fill between[of=LoStatVol and HiStatVol];
\addplot[thick, cStatVol] table[x=recall, y=P_StaticVolume, col sep=comma]{data/mean_pr_curves_for_latex.csv};

\addplot[name path=HiSelf, draw=none] table[x=recall, y=Hi_SelfAdaptive, col sep=comma]{data/mean_pr_curves_for_latex.csv};
\addplot[name path=LoSelf, draw=none] table[x=recall, y=Lo_SelfAdaptive, col sep=comma]{data/mean_pr_curves_for_latex.csv};
\addplot[fill=cSelf, fill opacity=0.12, draw=none] fill between[of=LoSelf and HiSelf];
\addplot[thick, cSelf] table[x=recall, y=P_SelfAdaptive, col sep=comma]{data/mean_pr_curves_for_latex.csv};

\addplot[name path=HiRF, draw=none] table[x=recall, y=Hi_RF, col sep=comma]{data/mean_pr_curves_for_latex.csv};
\addplot[name path=LoRF, draw=none] table[x=recall, y=Lo_RF, col sep=comma]{data/mean_pr_curves_for_latex.csv};
\addplot[fill=cRF, fill opacity=0.12, draw=none] fill between[of=LoRF and HiRF];
\addplot[thick, cRF] table[x=recall, y=P_RF, col sep=comma]{data/mean_pr_curves_for_latex.csv};

\addplot[name path=HiXGB, draw=none] table[x=recall, y=Hi_XGB, col sep=comma]{data/mean_pr_curves_for_latex.csv};
\addplot[name path=LoXGB, draw=none] table[x=recall, y=Lo_XGB, col sep=comma]{data/mean_pr_curves_for_latex.csv};
\addplot[fill=cXGB, fill opacity=0.08, draw=none] fill between[of=LoXGB and HiXGB];
\addplot[thick, cXGB] table[x=recall, y=P_XGB, col sep=comma]{data/mean_pr_curves_for_latex.csv};

\end{axis}
\end{tikzpicture}%

    \end{minipage}}
    
    \caption{\label{fig:f1_pr_compare} Classification performance for all methods under consideration. (a) F1-score, and (b) mean precision-recall curves for $\mathrm{Recall} \in [0.5, 1]$.}
\end{figure*}
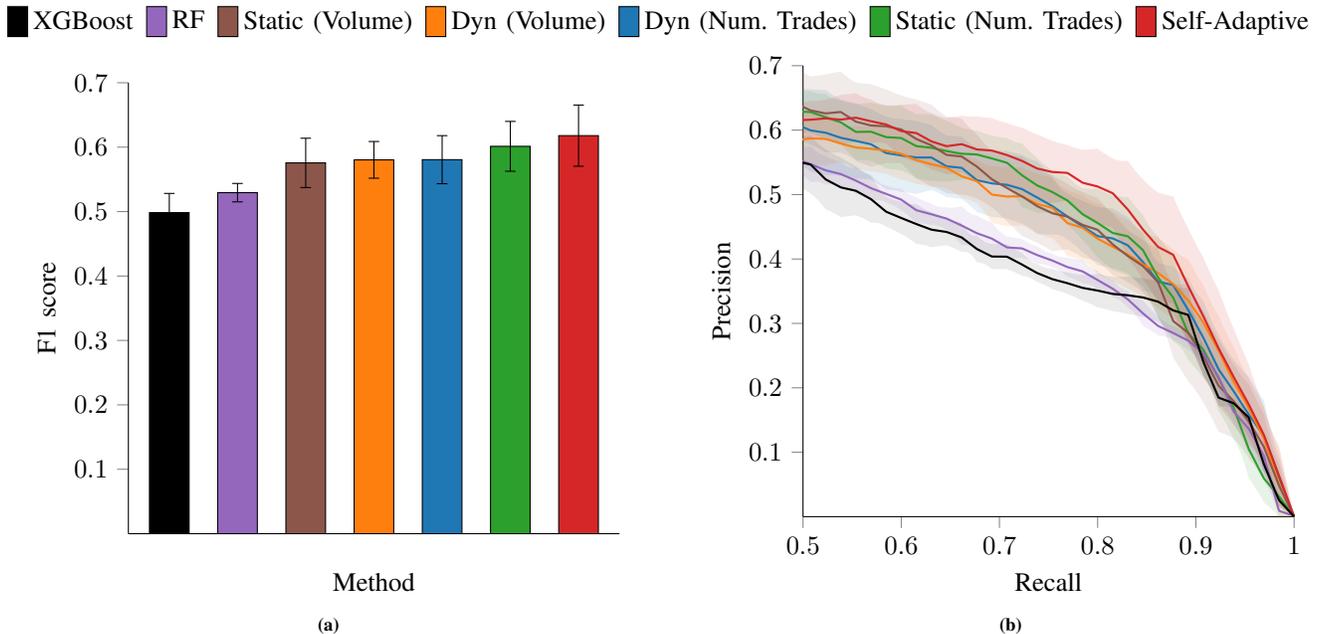


We compute $18$ additional features on this base panel to capture abrupt regime shifts, typical of coordinated manipulation. Since our dataset is limited to candle-level aggregates and does not include order-book information (e.g., bid--ask depth or spread), we construct a proxy for aggressive buy pressure using the ratio between taker buy quote volume and total traded volume at the candle level. For each token, we compute rolling statistics over a fixed window of $12$ hours and then consider their percentage change over time, following the feature construction in~\cite{LaMorgia2020}. Such percentage changes of rolling means or rolling standard deviations quantify how quickly the local average or local variability is shifting, which is the type of signal expected under coordinated activity. Following this approach, we compute percentage changes of rolling mean and standard deviation for the buy pressure, traded volume, number of trades, and close price. We then compute the same transformation for the rolling mean of the intraperiod maximum price. Last, since intraday patterns can be relevant when campaign timing is systematic, we derive an hour feature as the integer hour extracted from the timestamp. The full list of the features considered is offered in Table~\ref{tab:features_hourly} in Section~\ref{sec:Appendix} of the Appendix.

To respect temporal ordering and emulate a surveillance setting, we adopt a chronological hold-out split for all considered models. Let $\mathcal{T}$ denote the ordered set of unique hourly timestamps in the panel. We partition $\mathcal{T}$ into $60\%$ for training, $20\%$ for validation, and  $20\%$ for testing. Graph construction using historical observations is performed using only the training data, model hyperparameters and early stopping criteria are selected on the validation set. The decision threshold $\gamma$ is also selected on the validation set by maximizing the F1-score, and the same threshold is applied unchanged to the test set, which is utilized only for the final accuracy estimation. Then, to mitigate
short-range temporal dependence and reduce leakage around split boundaries, an embargo of $z = 5$ hours is adopted and the first $z$ timestamps of the validation and test blocks are
discarded~\cite{LopezDePrado2018}. We evaluate performance in terms of precision, recall, and F1 score, defined as:
\begin{equation}
    \label{eq:F1-score}
    \text{F1-score} = 2 \cdot \frac{\mathrm{Precision} \cdot \mathrm{Recall}}{\mathrm{Precision} + \mathrm{Recall}},
\end{equation}
where precision is defined as $\mathrm{Precision} =
\mathrm{TP}/(\mathrm{TP} + \mathrm{FP})$ and recall as $\mathrm{Recall} = \mathrm{TP}/(\mathrm{TP} + \mathrm{FN})$, with $\mathrm{TP}$, $\mathrm{FP}$, and $\mathrm{FN}$ denoting true positives, false positives, and false negatives classifications, respectively. All the results reported hereby correspond to their mean value after 9 runs.
For all experiments, we employ the ST-GNN architecture defined in Section~\ref{sec:stgnn_arch} with a fixed learning rate of $10^{-3}$, and a spatial embedding dimension of $D=64$ (Eq.~\ref{eq:GTransformer}), $2$ attention heads, and a temporal window of $W=5$ hours. Regarding graph construction, for the correlation-based strategies (G1, G2), we set the minimum threshold to $\tau_{\min}=0.15$, the dynamic lookback window to $L=12$. The quantile density in (G1) is set to 
$\rho = 0.90$ when considering the number of trades, to $0.75$ when considering the volume, and to $0.95$ for both dynamic variants in (G1). Finally, for the self adaptive graph (G3), the learnable embedding dimension is $d=48$ and the sparsity threshold is $\epsilon=0.005$. The complete grid search spaces and selected configurations for all graph inference strategies are reported in Table~\ref{tab:graph_hyperparameters} in Appendix~\ref{sec:appendix_hyperparameters}.
All the results reported hereby correspond to their mean value after 9 runs.

\begin{figure*}[!t]
	\centering
	\begin{minipage}{\textwidth}
	\centering
	\hspace{+1.5cm}%
  \begin{tikzpicture}
  \centering
  \begin{customlegend}[
      legend columns=7,
      legend style={
        anchor=north,
        draw=none,
        fill=none,
        /tikz/every even column/.append style={column sep=0.10cm}},
      legend entries={XGBoost\\RF\\Static (Volume)
      \\Dyn (Volume) \\Dyn (Num. Trades) \\ Static (Num. Trades)
      \\ Self-Adaptive \\},
      legend image post style={xscale=0.8},
  ]
    \addlegendimage{area legend, fill=cXGB,       draw=black, legend image post style={xscale=0.55,yscale=2.0}}
    \addlegendimage{area legend, fill=cRF,        draw=black, legend image post style={xscale=0.55,yscale=2.0}}
    \addlegendimage{area legend, fill=cStatVol,   draw=black, legend image post style={xscale=0.55,yscale=2.0}}
    \addlegendimage{area legend, fill=cDynVol,    draw=black, legend image post style={xscale=0.55,yscale=2.0}}
    \addlegendimage{area legend, fill=cDynTrades, draw=black, legend image post style={xscale=0.55,yscale=2.0}}
    \addlegendimage{area legend, fill=cStatTrades,draw=black, legend image post style={xscale=0.55,yscale=2.0}}
    \addlegendimage{area legend, fill=cSelf,      draw=black, legend image post style={xscale=0.55,yscale=2.0}}
  \end{customlegend}
\end{tikzpicture}%

	\end{minipage}
	\vspace{0.8em}
%
  \begin{tikzpicture}
\begin{axis}[
    width=0.92\textwidth,
    height=0.34\textwidth,
    ybar,
    bar width=16pt,
    ymin=0, ymax=1,
    ytick={0,.2,.4,.6,.8,1.0},
    ylabel={F1 score},
    symbolic x coords={APPC,NXS},
    xtick={APPC,NXS},
    xticklabel style={rotate=45, anchor=east},
    enlarge x limits=0.48,
    tick align=outside,
    axis x line*=bottom,
    axis y line*=left,
]

\addplot [fill=cXGB, draw=black, error bars/.cd, y dir=both, y explicit]
coordinates {
    (APPC, .654) +- (.048,.048)
    (NXS,  .779) +- (.053,.053)
};

\addplot [fill=cRF, draw=black, error bars/.cd, y dir=both, y explicit]
coordinates {
    (APPC, .708) +- (.046,.046)
    (NXS,  .851) +- (.046,.046)
};

\addplot [fill=cStatVol, draw=black, error bars/.cd, y dir=both, y explicit]
coordinates {
    (APPC, .794) +- (.075,.075)
    (NXS,  .909) +- (.066,.066)
};

\addplot [fill=cDynVol, draw=black, error bars/.cd, y dir=both, y explicit]
coordinates {
    (APPC, .725) +- (.119,.119)
    (NXS,  .873) +- (.047,.047)
};

\addplot [fill=cDynTrades, draw=black, error bars/.cd, y dir=both, y explicit]
coordinates {
    (APPC, .741) +- (.103,.103)
    (NXS,  .903) +- (.088,.088)
};

\addplot [fill=cStatTrades, draw=black, error bars/.cd, y dir=both, y explicit]
coordinates {
    (APPC, .755) +- (.075,.075)
    (NXS,  .901) +- (.053,.053)
};

\addplot [fill=cSelf, draw=black, error bars/.cd, y dir=both, y explicit]
coordinates {
    (APPC, .808) +- (.050,.050)
    (NXS,  .931) +- (.057,.057)
};

\end{axis}
\end{tikzpicture}%

	\caption{\label{fig:token_results} Classification performance in terms of F1-score for all methods under consideration for tokens (i.e., APPC and NXS) with at least five pump events in the test set.}
\end{figure*}
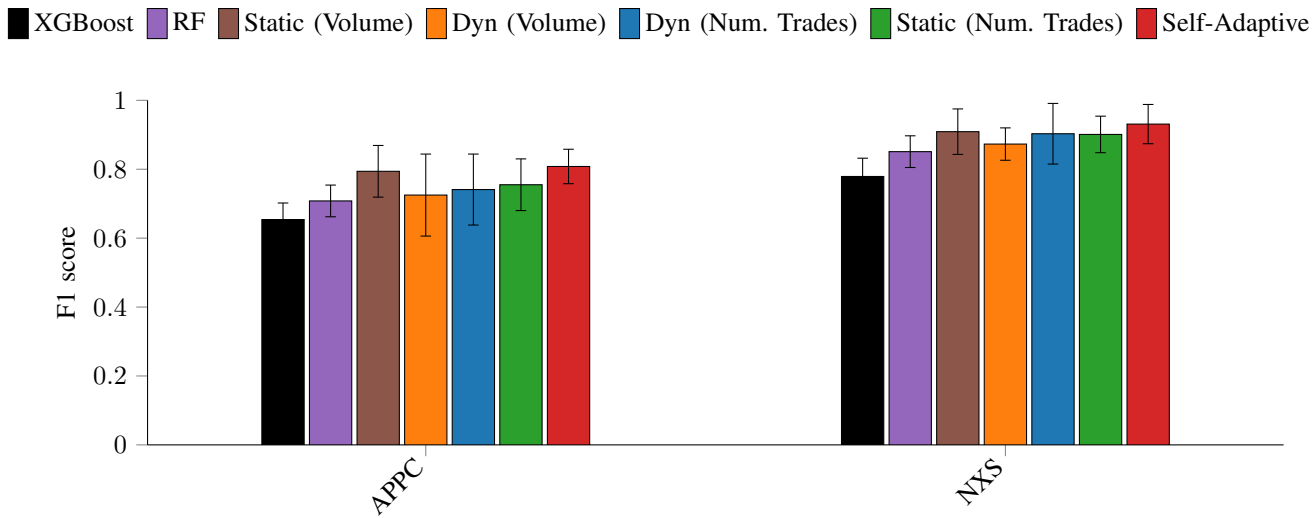

\subsection{Classification performance}
\label{sec:Res_ACC}
We present in Figure~\ref{fig:F1Bars} the F1-score~\eqref{eq:F1-score} of the classification of the minority (i.e., fraudulent) class. The two tree-based methods achieve the lowest scores, with XGBoost (in black) yielding the lowest $\mathrm{F1} = 0.49 \pm 0.03$ and Random Forests (in purple) improving upon it with $\mathrm{F1} = 0.53 \pm 0.01$. All our introduced graph-based variants outperform these baselines in terms of classification accuracy. The static graph based on volume correlation (in brown), and its dynamic counterpart (in orange) achieve $\mathrm{F1} = 0.575 \pm 0.04$ and
$\mathrm{F1} = 0.58 \pm 0.02$, respectively. The  strongest correlation-based performance is obtained when connectivity is derived from the number of trades as a proxy for trading activity 
with $\mathrm{F1} =  0.58 \pm 0.04$ for the dynamic graph (in blue) and $\mathrm{F1} = 0.60 \pm 0.04)$ for the static one (in green). The self-adaptive graph (in red) achieves the best overall accuracy with $F1 = 0.62 \pm 0.05$, indicating that allowing the model to adjust  connectivity during training is beneficial when the graph topology is not available. The moderate error bars across seeds confirm that these rankings are stable and not driven by favorable initializations. To further validate that the observed gains stem from the graph structure rather than from increased model capacity, we evaluate temporal-only baselines (GRU, TransformerEncoder) that process each token independently; full details are provided in Appendix~\ref{sec:appendix_temporal_baselines}. A comparison of model sizes and training times is reported in Table~\ref{tab:runtime_params} in Appendix~\ref{sec:appendix_runtime}.


In Figure~\ref{fig:PRCurve}, we report the mean Precision-Recall (PR) curves for $\mathrm{Recall \in [0.5, 1]}$. Given the extreme class imbalance in our test set, containing only $\sim 0.016\%$ positive pump hours in a total of $\sim 1.9 \times 10^6$ hours, we prioritize PR curves over ROC analysis, as the latter can be insensitive to the false-positive burden when negatives dominate \cite{saito2015pr}. The plot highlights a consistent performance gap, with graph-based models shifting the PR frontier upward relative to the tree-based baselines, implying  higher precision for any given recall level. While performance naturally degrades as recall approaches $1$, our graph-based methods maintain precision levels significantly longer than the tree-based baselines. The best method is the self-adaptive variant, which dominates the frontier across the majority of the sensitivity range. 

In Figure~\ref{fig:token_results} we present the classification performance in terms of
F1-score~\eqref{eq:F1-score} for the most affected tokens, with at least five recorded pump events 
in the test set. All graph-based variants introduced in Subection~\ref{sec:graph_inference} report
accuracy scores higher than the competing tree-based methods. The self-adaptive variant (in red) achieves the highest scores with $\mathrm{F1} = 0.808 \pm 0.05$ for APPC and $\mathrm{F1} = 0.931 \pm 0.06$ for NXS, where APPC and NXS are individual token ticker symbols, each representing a single node in the graph.

\section{Conclusions}
\label{sec:concl}
We presented a framework to detect manipulation in cryptocurrency markets,
specifically addressing the scenario where only aggregated market data are available, i.e.,
OHLCV and trading activity, while transaction-level information remains
unobservable. To overcome this limitation, we proposed a methodology that infers token-to-token
connectivity from market dynamics, and processes it via a unified spatio-temporal GNN (ST-GNN)
architecture. We proposed three graph inference methods: static correlation, event-driven dynamic correlation, and self-adaptive adjacency learning.

We compared our graph-based solutions against tree-based methods on a real-world labeled
dataset of pump-and-dump schemes spanning the period 2017--2021. Our results demonstrate that graph-based
models are superior in terms of F1-score and Precision-Recall
efficiency. Notably, the self-adaptive adjacency method achieved the highest fraud detection
performance, confirming that the model can successfully learn latent interaction structure between
tokens.
Our work highlights the effectiveness of the introduced graph-based fraud detection algorithms,
and their applicability for the monitoring of market manipulation events.

\section*{Acknowledgment}
We acknowledge the financial supported by the joint DFG-470857344
and SNSF-204817 project entitled ``Numerical Algorithms, Frameworks, and
Scalable Technologies for Extreme-Scale Computing'', and by the Huawei
Zurich Research Center. This work was also funded by the ``Resilient
Financial Enterprises'' project, supported by the Swiss Innovation Agency
(Innosuisse) under grant agreement 119.321 INT-ICT.

\bibliographystyle{IEEEtran} 
\bibliography{SDS26_BIB}        
\newpage
\section{Appendix}
\label{sec:Appendix}
\subsection{Features}
\begin{table}[h]
\centering
\small
\setlength{\tabcolsep}{4pt}
\begin{tabular}{llp{4.9cm}}
\toprule
Feature & Type & Definition \\
\midrule
OHLCV & raw & Open/high/low/close and volume. \\
quote\_asset\_volume & raw & Volume in quote currency. \\
num\_trades & raw & Trades per hour. \\
taker\_buy\_base & raw & Hourly taker-buy volume (base). \\
taker\_buy\_quote & raw & Hourly taker-buy volume (quote). \\
std\_rush\_order & eng & Pct. change 12h roll std (buy pressure). \\
avg\_rush\_order & eng & Pct. change 12h roll mean (buy pressure). \\
std\_trades & eng & Pct. change 12h roll std (num\_trades). \\
std\_volume & eng & Pct. change 12h roll std (volume). \\
std\_price & eng & Pct. change 12h roll std (close). \\
avg\_volume & eng & Pct. change 12h roll mean (volume). \\
avg\_price & eng & Pct. change 12h roll mean (close). \\
avg\_price\_max & eng & Pct. change 12h roll mean (high). \\
hour\_of\_the\_day & time & UTC hour of the day. \\
\bottomrule
\end{tabular}

\caption{Feature set used in the numerical experiment. Each token corresponds to a graph node $i\in\mathcal{V}$, and the features listed here constitute the node feature vector $\mathbf{x}_{i,t}\in\mathbb{R}^{F}$ at each hourly timestamp. Raw features are obtained directly from the market data downloader, engineered features (eng) are constructed during the feature engineering stage, and temporal features (time) are derived from the UTC timestamp of each observation.}
\label{tab:features_hourly}
\end{table}

\subsection{Hyperparameter selection for graph inference}
\label{sec:appendix_hyperparameters}

To isolate the effect of the graph inference strategy, the ST-GNN architecture was kept fixed across all experiments, and only the graph-inference hyperparameters and the dropout probability were tuned. For all methods, the learning rate was fixed to \(10^{-3}\). For the correlation-based graphs \((\mathrm{G1})\) and \((\mathrm{G2})\), the scalar series \(s_{i,t}\) used to construct the graph was chosen from a single raw feature, namely \texttt{num\_trades} or \texttt{volume}. The final selected configurations are reported in Table~\ref{tab:graph_hyperparameters}. The ST-GNN with \((\mathrm{G1})\) and \((\mathrm{G2})\) contains 135,041 learnable parameters, whereas
\((\mathrm{G3})\) contains 143,105 due to the additional learnable node embedding matrices
\(\mathbf{E}_1\) and \(\mathbf{E}_2\). Parameter counts are also summarized in
Table~\ref{tab:runtime_params}.

\begin{table}[h]
\centering
\footnotesize
\setlength{\tabcolsep}{4pt}
\renewcommand{\arraystretch}{1.2}
\begin{tabular}{@{} p{2.6cm} p{3.8cm} p{1.8cm} @{}}
\toprule
Signal & Search Space & Selected \\
\midrule

\multicolumn{3}{@{}l}{\textbf{(G1) Static}} \\[2pt]
$s_{i,t}=\texttt{num\_trades}$ & 
dropout $\in \{0.2,0.3,0.4\}$ \newline $\rho \in \{0.95,0.90,0.75\}$ & 
dropout $=0.3$ \newline $\rho=0.90$ \\[8pt]

$s_{i,t}=\texttt{volume}$ & 
dropout $\in \{0.2,0.3,0.4\}$ \newline $\rho \in \{0.95,0.90,0.75\}$ & 
dropout $=0.2$ \newline $\rho=0.75$ \\
\midrule

\multicolumn{3}{@{}l}{\textbf{(G2) Dynamic}} \\[2pt]
$s_{i,t}=\texttt{num\_trades}$ & 
dropout $\in \{0.2,0.3,0.4\}$ \newline $L \in \{12,24\}$ \newline $\rho \in \{0.95,0.90,0.75\}$ & 
dropout $=0.3$ \newline $L=12$ \newline $\rho=0.95$ \\[12pt]

$s_{i,t}=\texttt{volume}$ & 
dropout $\in \{0.2,0.3,0.4\}$ \newline $L \in \{12,24\}$ \newline $\rho \in \{0.95,0.90,0.75\}$ & 
dropout $=0.2$ \newline $L=12$ \newline $\rho=0.95$ \\
\midrule

\multicolumn{3}{@{}l}{\textbf{(G3) Self-Adaptive}} \\[2pt]
\textit{} & 
dropout $\in \{0.2,0.3,0.4\}$ \newline $d \in \{16,32,48,64\}$ \newline $\epsilon \in \{0.005,0.001,0.02\}$ & 
dropout $=0.4$ \newline $d=48$ \newline $\epsilon=0.005$ \\

\bottomrule
\end{tabular}
\caption{Grid search spaces and final selected hyperparameters for the three graph inference strategies. For \((\mathrm{G1})\) and \((\mathrm{G2})\), the scalar series \(s_{i,t}\) used to construct the correlation-based graphs is reported.}
\label{tab:graph_hyperparameters}
\end{table}

\subsection{Temporal-only baselines}
\label{sec:appendix_temporal_baselines}

To further assess the contribution of the inferred graph structure, we also evaluated two temporal-only
baselines, namely a GRU and a TransformerEncoder, which process token histories independently and
therefore cannot model cross-token coordination. Among them, the GRU was the strongest temporal-only
baseline, achieving a maximum F1-score of \(0.56\) and a PR-AUC of \(0.51\). This remains below the
best graph-based model, for which the self-adaptive graph \((\mathrm{G3})\) reaches a PR-AUC of \(0.58\).

Importantly, increasing the GRU capacity from approximately \(16\)k to \(164\)k trainable parameters did not close the gap, indicating that the improvement of the proposed approach does not arise merely from higher model capacity, but from the use of an inferred relational structure that enables cross-token information exchange.

\subsection{Computational footprint and training time}
\label{sec:appendix_runtime}

Table~\ref{tab:runtime_params} reports the approximate model size and training time of the main
baselines and the proposed graph-based variants. Overall, \((\mathrm{G1})\) Static provides the
best trade-off between performance and computational cost. Although \((\mathrm{G3})\) Self-Adaptive is the most expensive among the proposed graph-based models, it remains practically deployable and is comparable to stronger non-neural baselines in wall-clock time. All experiments were executed on a single NVIDIA RTX A6000 GPU, with each model trained in isolation (full GPU availability).
 
\begin{table}[h]
\centering
\small
\setlength{\tabcolsep}{4pt}
\renewcommand{\arraystretch}{1.1}
\begin{tabular}{@{} p{3.2cm} p{2.4cm} p{2.6cm} @{}}
\toprule
Model & Approx. Num. \newline Parameters & Approx. \newline Training Time \\
\midrule
GRU (small) & $\sim 16$k & $\sim 3$ min \\
GRU (large) & $\sim 164$k & $\sim 4$ min \\
TransformerEncoder & $\sim 64$k & $\sim 12$ min \\
(G1) Static & $\sim 135$k & $\sim 15$ min \\
(G2) Dynamic & $\sim 135$k & $\sim 31$ min \\
(G3) Self-Adaptive & $\sim 143$k & $\sim 58$ min \\
XGBoost & $\sim 15$k & $\sim 60$ min \\
Random Forest & $\sim 500$k & $\sim 60$ min \\
\bottomrule
\end{tabular}
\caption{Approximate model size and training time of the temporal-only baselines, graph-based variants, and classical baselines. Times are wall-clock approximations under the same experimental setup. For classical machine learning baselines, we report approximate model size rather than neural trainable parameters.}
\label{tab:runtime_params}
\end{table}


\end{document}